\documentclass{article}

\usepackage[final]{neurips_2025}

\usepackage[utf8]{inputenc} 
\usepackage[T1]{fontenc}    
\usepackage{hyperref}       
\usepackage{url}            
\usepackage{booktabs}       
\usepackage{amsfonts}       
\usepackage{nicefrac}       
\usepackage{microtype}      
\usepackage{xcolor}         
\usepackage{graphicx}
\usepackage{subcaption}
\usepackage{amsmath}

\usepackage{enumitem}
\setlist{itemsep=0pt, topsep=2pt, parsep=0pt, partopsep=0pt}

\makeatletter
\renewcommand{\@noticestring}{NeurIPS \@neuripsyear\ AI for Science Workshop.}
\makeatother

\title{From In Silico to In Vitro: Evaluating Molecule Generative Models for Hit Generation}
\workshoptitle{AI for Science Workshop (NeurIPS 2025)}

\author{%
  Nagham Osman\\
  University College London\\
  London, UK \\
  \texttt{nagham.osman.21@ucl.ac.uk} \\
  \And
  Vittorio Lembo \\
  University of Urbino Carlo Bo \\
  Urbino, Italy \\
  \AND
  Giovanni Bottegoni \\
  University of Urbino Carlo Bo \\
  Urbino, Italy \\
  \And
  Laura Toni \\
  University College London \\
  London, UK \\
}

\begin{document}

\maketitle

\begin{abstract}
Hit identification is a critical yet resource-intensive step in the drug discovery pipeline, traditionally relying on high-throughput screening of large compound libraries. Despite advancements in virtual screening, these methods remain time-consuming and costly. Recent progress in deep learning has enabled the development of generative models capable of learning complex molecular representations and generating novel compounds \textit{de novo}. However, using ML to replace the entire drug-discovery pipeline is highly challenging. In this work, we rather investigate whether generative models can replace one step of the pipeline: \textit{hit-like} molecule generation. To the best of our knowledge, this is the first study to explicitly frame hit-like molecule generation as a standalone task and empirically test whether generative models can directly support this stage of the drug discovery pipeline. Specifically, we investigate if such models can be trained to generate \textit{hit-like} molecules, enabling direct incorporation into, or even substitution of, traditional hit identification workflows. We propose an evaluation framework tailored to this task, integrating physicochemical, structural, and bioactivity-related criteria within a multi-stage filtering pipeline that defines the \textit{hit-like} chemical space. Two autoregressive and one diffusion-based generative models were benchmarked across various datasets and training settings, with outputs assessed using standard metrics and target-specific docking scores. Our results show that these models can generate valid, diverse, and biologically relevant compounds across multiple targets, with a few selected GSK-3$\beta$ hits synthesized and confirmed active in vitro. We also identify key limitations in current evaluation metrics and available training data.
\end{abstract}

\section{Introduction}
\label{sec: intro}
Traditionally, drug discovery begins with target validation, followed by hit identification which is the first stage introducing chemical matter and novelty. A \textit{hit} is a small molecule with reproducible activity, acceptable synthetic accessibility, and physicochemical properties~\citep{goodnow_hit_2006}. Identifying high-quality hits is critical, as it initiates the hit-to-lead process to improve potency, selectivity, and pharmacokinetic properties. To guide this process, medicinal chemists often rely on heuristics such as Lipinski’s Rule of Five, which defines drug-likeness based on molecular weight, lipophilicity, hydrogen-bonding capacity, and related efficiency metrics that balance potency with physicochemical properties. Despite its importance, hit identification remains a resource-intensive task. Various experimental strategies have been developed, including high-throughput screening (HTS) of libraries of commercially available compounds, in-house collections, or natural products, aimed at identifying biologically active hits. However, the main bottleneck lies in the fact that these approaches are both time-consuming (months to years)~\citep{paul_how_2010} and financially demanding~\citep{hughes_principles_2011}. To partially overcome these limitations, computational methods such as structure-based virtual screening have been adopted to efficiently prioritize promising molecules. Nevertheless, even with these tools, identifying high-quality hits from vast chemical libraries continues to represent a major bottleneck in drug discovery, as these methods also entail considerable time and resource demands.

In recent years, deep learning (DL) has gained interest in molecular design due to its potential to learn complex structures and property relationships from large chemical datasets. To address persistent challenges in early drug discovery for \textit{de novo} molecule generation, research has increasingly explored generative models that are trained either to replicate the distribution of known compounds or to optimize specific constraints, such as molecular properties or binding affinity. Common strategies include one-shot generation~\citep{de_cao_molgan_2018, vignac_digress_2022, zang_moflow_2020}, sequential atom- or bond-level construction~\citep{gebauer_inverse_2022, segler_generating_2018, zhou_optimization_2019}, and fragment-based assembly~\citep{podda_deep_2020, seo_fragment_2021, jin_multi_objective_2020, gupta_generative_2018}. However, prior research on generative models has predominantly focused on generating molecules in general, without considering how such models perform when applied to a specific step in the drug discovery process. This work investigates their applicability to a more specific and demanding task: generating hit molecules. It reframes the scientific question from the general inquiry: \textit{Can generative models produce valid, drug-like molecules?} to a more targeted one: \textit{Can generative models accelerate the hit identification process by directly generating compounds with hit-like characteristics?} Accordingly, our focus shifts from evaluating generative models based solely on chemical validity to assessing their ability to produce molecules suitable for entry into physics-based screening workflows. The ultimate goal is to identify biologically viable hits that can progress into hit-to-lead and lead optimization stages, either via specialized ML models or through conventional medicinal chemistry carried out by pharmaceutical experts. As the drug discovery process remains largely sequential and devoid of true shortcuts, we ask whether generative models can effectively replace one of its early steps, hit identification. If successful, this approach could substantially reduce the time and cost associated with early-stage drug discovery, potentially compressing a multi-year process into just a few months~\citep{zhavoronkov_deep_2019, xu_generative_2025}.

Several benchmarking and scoring frameworks for de novo molecular generation have been proposed in recent years~\citep{thomas_molscore_2024, cerveira_evaluation_2024} which provide flexible, target-aware scoring functions and standardized benchmarking protocols for drug-like molecule generation. These frameworks are primarily designed to compare optimization strategies or reward functions across diverse molecular design objectives.

In contrast, the main contribution of this work lies in defining, operationalizing, and empirically stress-testing hit generation as a realistic downstream task for molecular generative models. Our work focuses on a distinct and more narrowly defined task: hit generation as a standalone stage of the drug discovery pipeline. Rather than proposing new scoring functions or optimization objectives, we explicitly operationalize hit-likeness using medicinal-chemistry filters, distributional alignment to known ligands, and target-specific docking distributions, and evaluate whether off-the-shelf generative models can directly produce molecules suitable for downstream screening and experimental validation. To our knowledge, no prior work has reported a task-oriented, end-to-end evaluation of off-the-shelf generative models explicitly focused on hit generation, combining medicinal-chemistry constraints, target-specific docking distributions, and prospective in vitro validation.

To this aim, our contributions are the following: 
\begin{itemize}
    \item We define an evaluation framework tailored to the hit-generation task by combining structural, physicochemical, and bioactivity metrics with a new multi-stage filtering pipeline;
    \item We construct a comprehensive filtering pipeline by integrating established drug-likeness and medicinal chemistry constraints to define and evaluate the hit-like chemical space; 
    \item We benchmark representative models from autoregressive and diffusion-based paradigms across diverse datasets and training settings, and evaluate their outputs using standard and task-specific criteria, such as docking scores across a panel of biological targets;
    \item We demonstrate that generative models, when evaluated under explicit hit-like constraints, can produce not only chemically valid but also biologically meaningful hit candidates, including compounds that were prospectively validated through \textit{in vitro} assays against GSK-3$\beta$;    
    \item We identify key limitations in current generative models and publicly available datasets that lack training data that are explicitly tailored to early-stage hit identification.
\end{itemize}

Our results demonstrate that deep generative models, when trained and fine-tuned appropriately, can generate hit-like molecules suitable for early-stage screening, including compounds with confirmed in vitro activity. These findings highlight both the potential and current limitations of generative models in practical hit discovery. 

\section{Method: Benchmarks, Training, and Evaluation Pipelines}
\label{sec: method}

We evaluate three graph-based generative models, two autoregressive and one diffusion-based, that have shown strong performance in general molecule generation but have not been assessed for the specific task of hit generation. This section outlines the models, datasets, training settings, evaluation pipelines, and in vitro validation of GSK-3$\beta$ hits.

\subsection{Studied Models}
\label{sec: models}

The three graph-based models considered are MolRNN~\citep{li_multi-objective_2018}, GraphINVENT~\citep{mercado_graph_2020}, and DiGress~\citep{vignac_digress_2022}. MolRNN and GraphINVENT are autoregressive, sequentially building molecular graphs step by step. They differ in their graph encoders: MolRNN uses a gated recurrent unit (GRU), whereas GraphINVENT employs a message-passing neural network (MPNN). DiGress, in contrast, uses a diffusion-based, one-shot generation approach with a graph transformer. This selection was based on the models' performance in preliminary evaluations and allows comparison of sequential versus non-sequential generation strategies, and the impact of different encoder architectures (GRU, MPNN, transformer) on scalability, and chemical validity.

\paragraph{MolRNN}
MolRNN~\citep{li_multi-objective_2018} is a sequential generative model that constructs molecular graphs by leveraging graph neural networks (GNNs) and recurrent neural networks (RNNs). It starts with an empty graph and iteratively guides the generation process using a probabilistic decoding policy to apply one of three actions: adding atoms, connecting atoms, or terminating the process. Each step involves computing atom-level embeddings through stacked GNN layers and refining graph-level representations with a GRU unit. A multi-layer perceptron is then used to predict action probabilities. The model is trained via maximum likelihood estimation using importance sampling. Full technical details are available in~\citep{li_multi-objective_2018}.

\paragraph{GraphINVENT}
GraphINVENT~\citep{mercado_graph_2020} is a graph-based generative model that constructs molecules by autoregressively adding atoms and bonds. GraphINVENT’s model architecture is similar to that of MolRNN; however, it employs MPNNs, which offer improved scalability and more fine-grained edge handling compared to GRUs. Molecular generation is guided by an Action Probability Distribution (APD) that restricts to chemically valid actions. The architecture features an MPNN encoder and a readout block to produce APD logits, trained by minimizing KL divergence against ground-truth actions derived from BFS-based preprocessing. See~\citep{mercado_graph_2020} for a comprehensive description.

\paragraph{DiGress}
DiGress~\citep{vignac_digress_2022} is a discrete denoising diffusion model for graph generation. Instead of continuous noise, it introduces categorical perturbations to node and edge features via discrete Markov transitions to ensure discreteness of its representation. For the denoising phase, a graph transformer network is employed that learns to reverse the noise process, using cross-entropy loss over node and edge predictions. The model incorporates structural and spectral features, such as cycle counts and Laplacian eigenvalues, to enhance its generative capabilities. Full methodology and results are detailed in~\citep{vignac_digress_2022}.

\subsection{Datasets}
To train our models, we utilize three types of datasets in this work: a general-purpose dataset (REINVENT~\citep{blaschke_reinvent_2020}), a filtered subset we term the hit-like dataset, and a target-specific ligand dataset.

\paragraph{REINVENT Dataset}
The REINVENT dataset~\citep{olivecrona_molecular_2017} is a widely used subset of ChEMBL~\citep{mendez_chembl_2019}, originally introduced by Olivecrona et al. It contains 1,086,248 unique compounds after filtering for molecules with 10–50 heavy atoms and restricting elements to H, B, C, N, O, F, Si, P, S, Cl, Br, and I. Duplicate compounds were removed to ensure uniqueness.

\paragraph{Hit-like Dataset}
The hit-like dataset, constructed for this study, is derived from the same ChEMBL source but filtered using the hit-like criteria such as molecular weight, synthetic accessibility score, ring constraints, and others described in Sec.~\ref{sec: filters}. 
This process yielded 58,837 molecules, approximately 2.68\% of the original database, representing structures with desirable drug-like properties.

\paragraph{Target-Specific Ligand Sets}
\label{sec: targets}
While our main focus is unconditional generation, target-specific ligand sets were also constructed to explore whether such models could be refined toward activity against defined biological targets, providing a complementary perspective on their downstream applicability. To construct these datasets, we retrieved ligands from ChEMBL annotated with activity against a panel of seven protein targets: dopamine D3 receptor (D3R, class A GPCR), adenosine A2A receptor (ADORA2A, class A GPCR), heat shock protein HSP90$\alpha$ (HSP90$\alpha$, molecular chaperone), glycogen synthase kinase-3$\beta$ (GSK-3$\beta$, serine/threonine kinase), proto-oncogene tyrosine-protein kinase Src (SRC, tyrosine kinase), coagulation factor II (thrombin, protease), and peroxisome proliferator-activated receptor $\alpha$ (PPAR$\alpha$, nuclear receptor).
Ligands were selected based on two criteria: (i) pChEMBL value $\geq 5$, and (ii) assay confidence score $= 9$. 

\subsection{Training Pipeline}
\begin{figure}[t]
    \centering
    \includegraphics[width=0.8\linewidth]{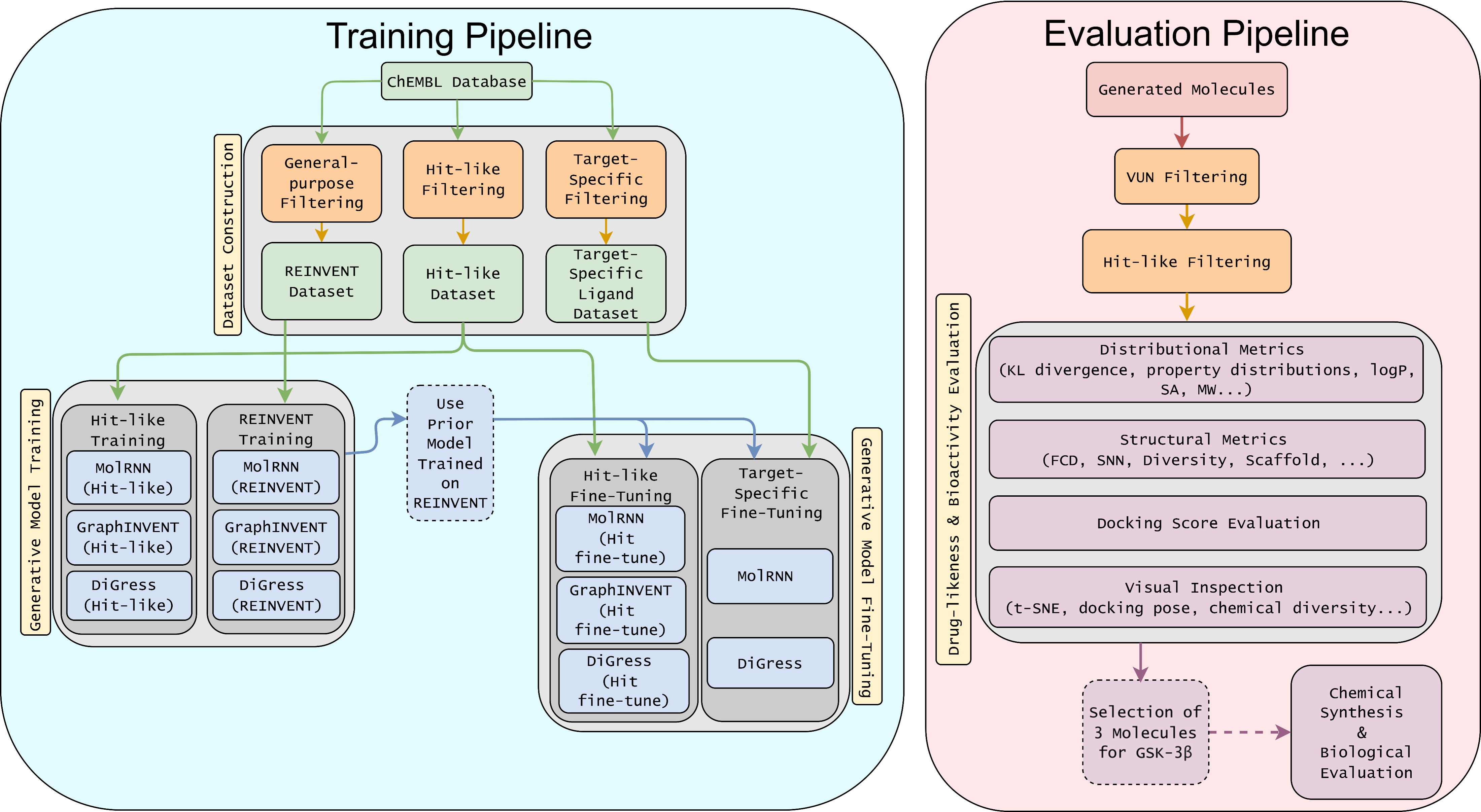} 
    \caption{\textit{Integrated Training and Evaluation Framework.} \textbf{Left:} Dataset construction and model training pipeline, showing the filtering strategies applied to ChEMBL to obtain general-purpose, hit-like, and target-specific datasets, their use in training MolRNN, GraphINVENT, and DiGress models, and subsequent fine-tuning in hit-like and target-specific settings. \textbf{Right:} Evaluation pipeline for generated molecules, outlining the multi-stage workflow from VUN and hit-like filtering through distributional, structural, and docking-based metrics, visual inspection, and final selection of top GSK-3$\beta$ candidates for synthesis and in vitro testing.}
    \label{fig:training-pipeline}
\end{figure}

As shown in the left hand side of Figure~\ref{fig:training-pipeline}, each model was trained under three settings: (1) using the full REINVENT dataset, (2) using the hit-like dataset, and (3) fine-tuning a REINVENT-trained model with the hit-like dataset. Additionally, MolRNN and DiGress models that were initially trained on the REINVENT dataset were further fine-tuned in a fourth setting on seven target-specific ligand sets (see \ref{sec: targets} for details). This choice was made to enable comparison between one autoregressive and one diffusion-based model. MolRNN was selected over GraphINVENT for this step due to its superior performance in preliminary benchmarks. 

All models followed their original training objectives, with early stopping on validation loss. For fine-tuning experiments, models were initialized from pretrained REINVENT weights and continued training for up to 20 epochs, using smaller batch sizes to avoid overfitting. For generation, the best-performing checkpoint for each trained model (lowest validation loss) was selected. To ensure fair comparison across models during downstream analysis, we continued generation until each model produced the same number of VUN and hit-like filtered molecules. This controlled sampling allowed consistent evaluation on docking and drug-likeness metrics. After training and sampling, each model was passed to the evaluation pipeline described in Sec.~\ref{sec:evaluation-pipeline}.


\subsection{Evaluation Pipeline}
\label{sec:evaluation-pipeline}
To  assess model performance, we adopt a three-layer evaluation framework of increasing drug relevance. Our goal is to construct a realistic evaluation pipeline (Figure~\ref{fig:training-pipeline}, right), aiming to determine whether generative models can replace hand-crafted hit-like molecule design.

\subsubsection{Basic Structural Validity Metrics}
The first layer of evaluation focuses on foundational checks to ensure that generated molecules are structurally and chemically sound. Validity measures the percentage of generated molecules that obey valence rules and contain no structural errors. Uniqueness assesses the proportion of valid molecules that are structurally distinct within the generated set. Novelty quantifies the fraction of valid and unique molecules that are not present in the training dataset. These three metrics (VUN) serve as the minimal baseline for generative model performance and are standard across the field.

\subsubsection{Distributional and Structural Similarity Metrics}
The second evaluation layer measures how closely generated molecules resemble the training chemical space. We employ a set of metrics that assess both molecular properties and structural patterns that align with standard benchmarking protocols such as GuacaMol~\citep{brown_guacamol_2019} and MOSES~\citep{polykovskiy_molecular_2018}. We use Fréchet ChemNet Distance (FCD) \cite{preuer_frechet_2018} for overall distributional similarity based on chemical and biological features of generated and reference molecules, BRICS fragments~\citep{degen_art_2008} and Bemis–Murcko scaffolds~\citep{bemis_properties_1996} similarities for structural patterns, Similarity to Nearest Neighbor (SNN) for proximity to known compounds based on Tanimoto similarity, and Jaccard-based internal diversity on Morgan fingerprints~\citep{morgan_generation_1965} for variation within the generated set.

\subsubsection{Drug-likeness and Bioactivity-Relevant Metrics}
This final evaluation tier assesses whether generative models produce molecules with realistic hit-like properties and potential biological activity. This aspect is one of the key contributions of our work where we emphasize evaluation by drug-likeness and biological relevance which are critical factors in determining the practical value of generated compounds.

\paragraph{Hit-like Filters}
\label{sec: filters}
To align with early-stage discovery goals, we applied a multi-stage filtering pipeline integrating established drug-likeness and medicinal chemistry criteria. Each molecule was required to satisfy: (1) a Novartis-inspired~\citep{schuffenhauer_evolution_2020} severity score (Sev.) $\leq$ 10, which penalizes structural liabilities such as PAINS motifs~\citep{baell_new_2010}, reactive groups, toxicophores, and unstable chemotypes; (2) molecular weight (MW) between 150--350~Da; (3) logP between 1--3 as a proxy for solubility; (4) a minimum pChEMBL value of 5 at any known target to ensure meaningful bioactivity; (5) a synthetic accessibility score (SAS) $\leq$ 5; (6) between one (NoR) and four rings ($<4$R), with no ring exceeding eight atoms ($<8$t), a maximum of two rings larger than six atoms ($<2$R6t), and exclusion of fused (FusedR) or small aromatic (AromR) ring systems; and (7) elemental composition restricted to C, N, O, F, P, S, Cl, Br, and I. This integrated approach provides a consistent and stringent definition of hit-likeness across datasets and generative outputs.

\paragraph{Docking Calculation}  
To evaluate the biological relevance of the generated molecules, we computed docking scores estimating their binding affinity to the seven protein targets described in Sec.~\ref{sec: targets}. Docking simulations were performed only on molecules that satisfied both the VUN criteria and the complete set of hit-like filters, ensuring that only the most plausible candidates were assessed for target binding. Docking scores, with lower values indicating stronger predicted interactions, serve as an approximate measure of bioactivity and complement chemical and structural evaluations. Details of protein and ligand preparation, docking settings, and software are provided in Appendix~\ref{app: docking-studies}.

\paragraph{Biological Evaluation}
To assess biological relevance of generated molecules, we performed hit selection for GSK-3$\beta$, a target where our models showed low KL divergence and many compounds outperforming the median docking score of known ligands. Candidate molecules were filtered to meet two criteria: (1) structural novelty (Tanimoto distance $\geq 0.5$ from known binders), and (2) strong predicted binding (docking scores $<$ mean $-2$~SD). From this subset, compounds were prioritized based on strong docking scores, visual inspection of binding poses, and synthetic feasibility. This process led to the selection of three top compounds subjected to further experimental testing in vitro to validate their potency. This procedure reflects a realistic medicinal-chemistry triage rather than a retrospective enrichment analysis.

\section{Experimental Results and Discussions}
This section presents our experimental results, including general-purpose molecule generation, docking score calculations, and the biological evaluation of selected hits against the GSK-3$\beta$ target.

\begin{table}[t]
  \caption{Molecular generation performance metrics for MolRNN, GraphINVENT, and DiGress across training settings.}
  \label{tab:gen-perf}
  \centering
  \resizebox{\linewidth}{!}{%
    \setlength{\tabcolsep}{5pt}
    \renewcommand{\arraystretch}{1.12}
    \begin{tabular}{lcccccccc}
      \toprule
      Model & Valid (\%) & VUN (\%) & Filters (\%) & FCD & Frag. & Scaff. & SNN & Div. \\
      \midrule
      MolRNN (Hit-like)       & 99.98 & 64.60 & \textbf{76.87} & 1.74 & 0.93 & \textbf{0.87} & \textbf{0.61} & 0.87 \\
      MolRNN (REINVENT)       & 99.98 & \textbf{96.74} & 11.46 & 2.05 & 0.92 & 0.79 & 0.50 & 0.87 \\
      MolRNN (Hit fine-tune)  & 99.91 & 94.64 & 67.57 & 1.91 & 0.92 & 0.84 & 0.52 & 0.87 \\
      GraphINVENT (Hit-like)  & \textbf{100.00} & \textbf{96.42} & 58.73 & 2.10 & \textbf{0.99} & 0.77 & 0.47 & \textbf{0.88} \\
      GraphINVENT (REINVENT)  & \textbf{100.00} & \textbf{96.98} & 12.85 & 1.70 & \textbf{0.99} & 0.78 & 0.48 & \textbf{0.88} \\
      GraphINVENT (Hit fine-tune) & \textbf{100.00} & 93.06 & 50.80 & \textbf{1.13} & \textbf{0.99} & 0.86 & 0.53 & \textbf{0.88} \\
      DiGress (Hit-like)      & 70.00 & 69.15 & 65.59 & 3.01 & \textbf{0.99} & 0.68 & 0.44 & \textbf{0.88} \\
      DiGress (REINVENT)      & 75.81 & 75.26 & 10.27 & 2.63 & \textbf{0.99} & 0.65 & 0.45 & \textbf{0.88} \\
      DiGress (Hit fine-tune) & 71.62 & 71.51 & 14.11 & 2.91 & 0.98 & 0.63 & 0.45 & \textbf{0.88} \\
      \bottomrule
    \end{tabular}%
  }
\end{table}

\begin{table}[t]
  \caption{Percentage of generated molecules failing each filtering criterion.}
  \label{tab:filter-fail}
  \centering
  \resizebox{\linewidth}{!}{%
    \setlength{\tabcolsep}{5pt}
    \renewcommand{\arraystretch}{1.12}
    \begin{tabular}{lccccccccccc}
      \toprule
      Model & Sev. & SAS & MW & logP & NoR & $<\!4$R & $<\!8$t & $<\!2$R6t & AromR & FusedR & All \\
      \midrule
      MolRNN (Hit-like)      & \textbf{0.80} & \textbf{0.30} & 9.97 & \textbf{14.39} & 0.16 & \textbf{0.27} & \textbf{0.37} & 0.03 & 0.03 & \textbf{0.02} & \textbf{23.13} \\
      MolRNN (REINVENT)      & 7.29 & 2.60 & 77.69 & 66.01 & 0.47 & 14.35 & 1.14 & 0.17 & 0.10 & 0.41 & 88.54 \\
      MolRNN (Hit fine-tune) & 1.96 & 0.37 & 15.14 & 19.63 & 0.25 & 0.40 & 0.26 & \textbf{0.01} & \textbf{0.02} & 0.06 & 32.43 \\
      GraphINVENT (Hit-like) & 7.51 & 6.59 & 11.26 & 21.90 & 1.85 & 1.49 & 8.45 & 1.75 & 1.69 & 1.12 & 41.27 \\
      GraphINVENT (REINVENT) & 9.06 & 8.59 & 73.25 & 63.59 & 1.43 & 16.30 & 8.98 & 3.28 & 1.23 & 1.80 & 87.15 \\
      GraphINVENT (Hit fine-tune) & 5.00 & 3.49 & 22.94 & 31.32 & 1.04 & 1.86 & 4.24 & 0.96 & 0.65 & 0.68 & 49.20 \\
      DiGress (Hit-like)     & 4.58 & 3.19 & \textbf{4.57} & 16.36 & 0.33 & 2.22 & 10.30 & 1.44 & 2.56 & 0.86 & 34.41 \\
      DiGress (REINVENT)     & 7.15 & 10.18 & 78.03 & 62.39 & 0.24 & 22.09 & 13.70 & 3.62 & 0.87 & 2.31 & 89.73 \\
      DiGress (Hit fine-tune)& 4.79 & 10.42 & 74.80 & 54.40 & \textbf{0.04} & 25.29 & 15.25 & 4.05 & 0.71 & 1.96 & 85.89 \\
      \bottomrule
    \end{tabular}%
  }
\end{table}

\subsection{Molecular Generation Performance and Filter Compliance}
The first step is to evaluate whether the models could reliably generate chemically valid, unique, and novel (VUN) molecules. As shown in Table~\ref{tab:gen-perf}, MolRNN and GraphINVENT achieved near-perfect validity ($>99.9\%$) across all settings, while DiGress reached only $70$--$76\%$. Uniqueness was uniformly high, but novelty varied: GraphINVENT consistently exceeded $93\%$, MolRNN was sensitive to training data (dropping to $64.6\%$ in the Hit-like regime), and DiGress remained moderate ($\sim70$--$75\%$) across all cases. We next assessed how many VUN molecules passed the hit-like property filters. Direct training on the Hit-like dataset yielded the highest compliance (MolRNN: $76.9\%$), while REINVENT-trained models passed at $<12\%$, underscoring the mismatch between broad chemical priors and focused hit-like requirements. Fine-tuning substantially improved MolRNN and GraphINVENT compliance, but DiGress saw only modest gains. Distributional metrics in Table~\ref{tab:gen-perf} highlight trade-offs between novelty and hit-like resemblance. GraphINVENT (Hit fine-tune) achieved the lowest FCD (1.13), indicating the closest match to the reference, while MolRNN showed the highest scaffold similarity, and DiGress exhibited generally higher FCD and scaffold deviation.

To pinpoint limiting factors, Table~\ref{tab:filter-fail} reports per-filter failure rates. MolRNN (Hit-like) showed minimal violations across all criteria, while REINVENT-trained MolRNN and DiGress frequently exceeded MW and logP thresholds. GraphINVENT tended toward higher ring-complexity and logP failures, partially mitigated by fine-tuning. DiGress, while rarely producing ringless molecules, overrepresented large and fused ring systems, with high MW violations persisting even after fine-tuning. DiGress consistently underperformed autoregressive models in validity and hit-like filter compliance. We attribute this primarily to the challenges of discrete diffusion in low-data, highly constrained regimes such as hit-like chemical space. Unlike autoregressive models, which enforce chemical validity at each generation step, diffusion-based approaches rely on global denoising and appear more sensitive to distributional sparsity and strict post hoc filters. This suggests that, in their current form, diffusion graph models may be less well suited to direct hit generation without additional constraint mechanisms. Overall, autoregressive models, when trained or fine-tuned on hit-like data, achieve the best balance between generative breadth and property compliance, whereas diffusion-based generation shows lower baseline validity and weaker adaptation to narrow physicochemical constraints.

\subsection{Docking Analysis}
\begin{table}[t]
  \centering
  \caption{Average KL divergence (mean $\pm$ standard deviation) between the docking score distributions of generated molecules and the reference ligand sets across all targets: 
  (a) grouped by combination of algorithm and training set; 
  (b) grouped by algorithm.}
  \label{tab:kl}
  \begin{subtable}[t]{0.62\linewidth}
    \centering
      \begin{tabular}{lc}
        \toprule
        Model & KL divergence \\
        \midrule
        DiGress (REINVENT)       & $0.099 \pm 0.061$ \\
        DiGress (Hit-like)       & $0.009 \pm 0.012$ \\
        DiGress (Hit fine-tune)  & $0.089 \pm 0.056$ \\
        GraphINVENT (REINVENT)   & $0.100 \pm 0.067$ \\
        GraphINVENT (Hit-like)   & $0.003 \pm 0.002$ \\
        GraphINVENT (Hit fine-tune) & $0.087 \pm 0.062$ \\
        MolRNN (REINVENT)        & $0.083 \pm 0.056$ \\
        MolRNN (Hit-like)        & $0.004 \pm 0.003$ \\
        MolRNN (Hit fine-tune)   & $0.095 \pm 0.064$ \\
        \bottomrule
      \end{tabular}%
    \subcaption{By model and training set}
    \label{tab:kl-by-model}
  \end{subtable}%
  \hfill
  \begin{subtable}[t]{0.38\linewidth}
    \centering
      \begin{tabular}{lc}
        \toprule
        Algorithm & KL divergence \\
        \midrule
        DiGress      & $0.066 \pm 0.062$ \\
        GraphINVENT  & $0.064 \pm 0.067$ \\
        MolRNN       & $0.061 \pm 0.062$ \\
        \bottomrule
      \end{tabular}%
    \subcaption{By algorithm}
    \label{tab:kl-by-algorithm}
  \end{subtable}
\end{table}

Docking performance was evaluated across seven targets (Sec.~\ref{sec: targets}) using KL divergence, quantifying the similarity between docking score distributions of generated molecules and known active ligands for each target. As shown in Table~\ref{tab:kl-by-model}, models trained directly on the Hit-like set achieved the lowest divergences (all $<0.01$), reflecting strong physicochemical alignment with the reference ligands. Unfiltered REINVENT-trained models exhibited higher divergence ($0.083$–$0.100$), consistent with the broader and less constrained chemical space. Fine-tuned models achieved intermediate values, indicating partial adaptation toward the hit-like profile.

When averaged across targets, MolRNN achieved lowest KL divergence ($0.061\pm0.062$), with GraphINVENT and DiGress performing similarly (Table~\ref{tab:kl-by-algorithm}). Per-target analysis in Appendix~\ref{app:per-target-kl} revealed highest divergences for SRC and PPAR$\alpha$, reflecting their higher MW and logP of their known ligands compared to the generated set. In contrast, GSK-3$\beta$ and ADORA2A showed the closest alignment with the hit-like constraints.

While KL divergence captures distributional similarity, it does not assess binding efficacy. We therefore compared docking scores of generated compounds to known ligands for each target, calculating the fraction outperforming (lower) the ligand median (Table~\ref{tab:docking-scores}). Across 63 model--training combinations, HSP90$\alpha$ showed the highest success rates (38--52\%), PPAR$\alpha$ the lowest ($<1$\%), and GSK-3$\beta$ (moderate 10--16\%), with similar trends for D3R and SRC. Given that many ChEMBL ligands are optimized molecules occupying broader chemical space, we repeated the analysis using only ligands passing hit-like filters (see full table details in Table~\ref{tab:medians-full-vs-hitlike} in Appendix~\ref{app:per-target-kl}. This raised success rates for five targets, most notably for Thrombin ($<2$\% to 35--45\%), while GSK-3$\beta$ and HSP90$\alpha$ observed modest drops. MolRNN (Hit fine-tune) achieved the highest success in 6 of 7 targets, with DiGress (Hit fine-tune) narrowly leading for PPAR$\alpha$. The weakest performances came from DiGress (Hit-like) in four targets, GraphINVENT (REINVENT) in two, and GraphINVENT (Hit-like) for HSP90$\alpha$. Overall, fine-tuned models on hit-like data generated a substantial fraction of compounds with docking scores comparable to or better than known ligands, though performance varied by target.

\begin{table*}[t]
  \centering
  \caption{Median docking score values of the ligand sets for PPAR$\alpha$, HSP90$\alpha$, and GSK-3$\beta$, and the percentage of generated molecules achieving a lower (better) docking score than the respective median ligand score, grouped by generative model.}
  \label{tab:docking-scores}
  \footnotesize 
  \setlength{\tabcolsep}{3pt}
  \renewcommand{\arraystretch}{1.1}

  \begin{subtable}[t]{0.3\linewidth}
    \centering
\resizebox{1.1\linewidth}{!}{%
    \begin{tabular}{l l r r}
      \toprule
      Model & Target & Median & \% Better \\
      \midrule
      Hit-like & PPAR$\alpha$  & $-9.15$ & 0.38 \\
               & HSP90$\alpha$ & $-6.03$ & 39.97 \\
               & GSK-3$\beta$  & $-7.27$ & 10.13 \\
      \midrule
      REINVENT & PPAR$\alpha$  & $-9.15$ & 0.35 \\
               & HSP90$\alpha$ & $-6.03$ & 42.69 \\
               & GSK-3$\beta$  & $-7.27$ & 12.18 \\
      \midrule
      Hit fine-tune & PPAR$\alpha$  & $-9.15$ & 0.54 \\
                    & HSP90$\alpha$ & $-6.03$ & 40.22 \\
                    & GSK-3$\beta$  & $-7.27$ & 11.60 \\
      \bottomrule
    \end{tabular}
    }
    \subcaption{DiGress}
  \end{subtable}%
  \hfill
  \begin{subtable}[t]{0.3\linewidth}
    \centering
\resizebox{1.1\linewidth}{!}{%
    \begin{tabular}{l l r r}
      \toprule
      Model & Target & Median & \% Better \\
      \midrule
      Hit-like & PPAR$\alpha$  & $-9.15$ & 0.38 \\
               & HSP90$\alpha$ & $-6.03$ & 42.07 \\
               & GSK-3$\beta$  & $-7.27$ & 15.58 \\
      \midrule
      REINVENT & PPAR$\alpha$  & $-9.15$ & 0.46 \\
               & HSP90$\alpha$ & $-6.03$ & 43.05 \\
               & GSK-3$\beta$  & $-7.27$ & 15.60 \\
      \midrule
      Hit fine-tune & PPAR$\alpha$  & $-9.15$ & 0.52 \\
                    & HSP90$\alpha$ & $-6.03$ & 51.82 \\
                    & GSK-3$\beta$  & $-7.27$ & 20.59 \\
      \bottomrule
    \end{tabular}
    }
    \subcaption{MolRNN}
  \end{subtable}%
  \hfill
  \begin{subtable}[t]{0.3\linewidth}
    \centering
\resizebox{1.1\linewidth}{!}{%
    \begin{tabular}{l l r r}
      \toprule
      Model & Target & Median & \% Better \\
      \midrule
      Hit-like & PPAR$\alpha$  & $-9.15$ & 0.38 \\
               & HSP90$\alpha$ & $-6.03$ & 38.35 \\
               & GSK-3$\beta$  & $-7.27$ & 14.00 \\
      \midrule
      REINVENT & PPAR$\alpha$  & $-9.15$ & 0.28 \\
               & HSP90$\alpha$ & $-6.03$ & 39.78 \\
               & GSK-3$\beta$  & $-7.27$ & 13.38 \\
      \midrule
      Hit fine-tune & PPAR$\alpha$  & $-9.15$ & 0.34 \\
                    & HSP90$\alpha$ & $-6.03$ & 39.07 \\
                    & GSK-3$\beta$  & $-7.27$ & 14.86 \\
      \bottomrule
    \end{tabular}
    }
    \subcaption{GraphINVENT}
  \end{subtable}
\end{table*}

\subsection{Hit Selection and Biological Evaluation}
Given its strong generative performance, GSK-3$\beta$ was chosen for hit prioritization, yielding three candidate molecules that were synthetized and tested in vitro. All showed high structural similarity (TD $>0.5$) to known kinase inhibitors, consistent with the conserved nature of kinase binding pockets. Compound~1 displayed strong GSK-3$\beta$ inhibition (IC$_{50}$ = 314~$\pm$~15.3~nM; Table~\ref{tab:bio-assay} in Appendix~\ref{app: bio-eval}; Figure~\ref{fig:bio-activity}A), while compounds~2 and~3 showed weaker activity. Such low-nanomolar activity, as in compound~1, is highly promising at the hit stage, particularly for a small, structurally novel scaffold. Notably, the training dataset contained few ligands of comparable potency and size, making compound~1 a clear example of successful extrapolation by the generative model. Docking analysis confirmed the hinge-binding of compound~1 through two key hydrogen bonds (Asp133, Val135) and a Lys85 interaction, explaining its high potency despite the rarity of its scaffold ($<1.1\%$ among kinase binders present in ChEMBL).

As shown in Figure~\ref{fig:bio-activity}, chemical space analysis further supported its relevance: t-SNE projections revealed that compound~1 clusters with known GSK-3$\beta$ inhibitors, while activity benchmarking demonstrated that it outperforms both the full ligand set and the subset of hit-like inhibitors (--log(activity) = 6.50 vs.\ median values of 6.39 and 5.89, respectively). These results highlight that generative models can design chemically novel yet biologically potent compounds, even without explicit fine-tuning on a specific target, by leveraging kinase features embedded in the training data.

\begin{figure}[t]
    \centering
    \includegraphics[width=0.8\linewidth]{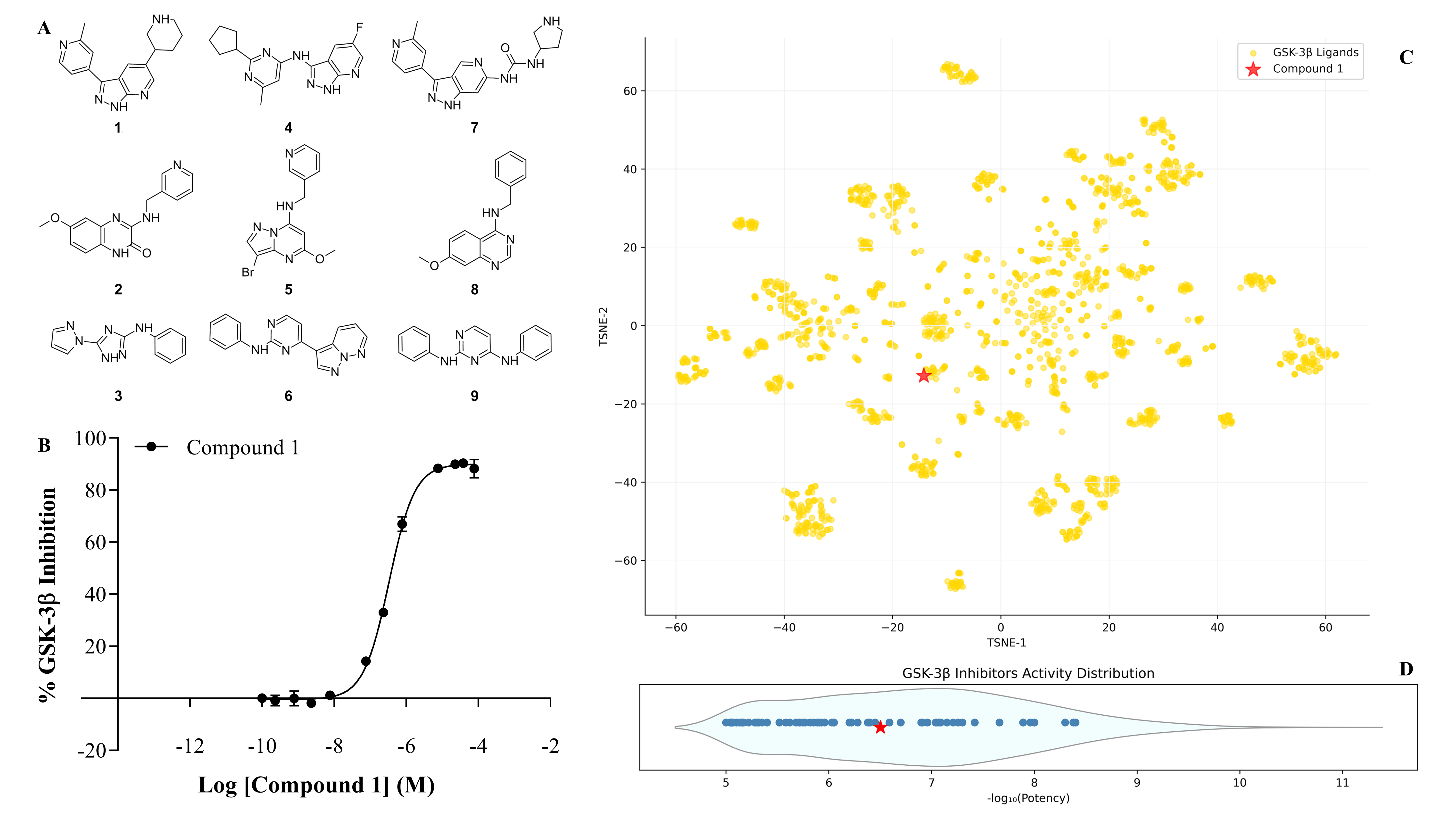} 
    \caption{(A) Structures of selected compounds (1--3), their most similar known GSK-3$\beta$ inhibitors (4--6), and closest kinase inhibitors (7--9) by TD. (B) Dose--response curve for GSK-3$\beta$ inhibition by compound~1, shown as mean $\pm$ SD of three replicates, representative of three experiments. (C) t-SNE projection of compound~1 (red star) within the chemical space of known GSK-3$\beta$ ligands (gold dots). (D) Comparison of pChEMBL value of compound~1 with known inhibitors, shown as a violin plot (all inhibitors, light blue) and subset with hit-like properties (blue dots).}
    \label{fig:bio-activity}
\end{figure}

\section{Conclusion}
This work presents a task-focused evaluation of generative models explicitly framed around hit generation, combining medicinal-chemistry filtering, docking-based distributional analysis, and prospective experimental in vitro validation. By isolating hit generation as a standalone objective, our study complements existing benchmarking and scoring frameworks by assessing whether generative models can directly produce candidates suitable for early-stage experimental screening. Our results demonstrate that such models can generate chemically novel compounds meeting hit-like criteria and displaying measurable biological activity, confirming their potential to accelerate early-stage drug discovery. Nonetheless, limitations remain. Diffusion-based models, despite strong performance in distribution learning, struggled with target-specific fine-tuning, due to their higher data demands and sensitivity to dataset size. Performance on targets such as PPAR$\alpha$ and SRC was constrained by the scarce high-quality hit-like data highlighting the broader issue of limited availability even for well-studied proteins. Moreover, widely used evaluation metrics such as VUN, FCD, and scaffold similarity did not consistently correlate with predicted bioactivity, underscoring the need for benchmarks that better capture biological relevance.

Future progress will require richer, better-curated hit-like datasets to support both general and target-specific training, as shown by constrained performance on PPAR$\alpha$ and SRC due to scarce ligands with suitable physicochemical properties. In parallel, evaluation frameworks must integrate bioactivity-relevant endpoints alongside distributional and structural metrics. Advances in architectures that learn effectively from limited target-specific data through improved transfer learning or data augmentation strategies will also be key. Incorporating these improvements into a modular AI-driven pipeline, where generative models focus on high-quality hit generation followed by ML-based or traditional hit-to-lead optimization, offers a promising path to accelerate early-stage drug discovery while addressing current gaps in predictive power and applicability.

\bibliographystyle{plainnat}
\bibliography{references}


\appendix
\section{Docking Studies Details}
\label{app: docking-studies}
Ligand docking calculations were carried out using Glide~\citep{halgren_glide_2004, friesner_glide_2004} as implemented in Schr\"odinger~2023--04 (Schr\"odinger LLC, New York, USA). All ligands were prepared using the LigPrep tool. For each compound, stereoisomers were generated for all chiral centers, up to a maximum of~32. Ionization states were generated using Epik Classic, considering a pH range of $7.4 \pm 1.0$. All other parameters were kept at their default settings. All receptors were prepared using the Protein Preparation tool implemented in Schr\"odinger~2023--04. Default settings were used. Water molecules, ions, and other non-protein atoms were removed from protein structures.

The position and size of the binding boxes of GSK-3$\beta$ (PDB code: 1Q41~\citep{bertrand_structural_2003}, chain~A), thrombin (PDB code: 2FZZ~\citep{pinto_aminobenzisoxazol_2006}, chain~H), and SRC (PDB code: 7OTE~\citep{cuesta-hernandez_allosteric_2023}, chain~A) systems were defined by the coordinates of their co-crystallized ligand. To account for receptor flexibility, the van der Waals radii for atoms with a partial charge $\leq 0.25$ were scaled to~0.9 for each system. For the HSP90$\alpha$ system (PDB code: 2VCI~\citep{brough_45-diarylisoxazole_2008}, chain~A), the binding box was defined in the same manner as described above, except that three water molecules near the key residue Asp93---reported to form a hydrogen-bond network between the ligand and the receptor---were retained. 

The binding boxes of the PPAR$\alpha$ system (PDB code: 6KB4~\citep{kamata_ppar_2020}, chain~A), D3R (PDB code: 3PBL~\citep{chien_structure_2010}, chain~A), and ADORA2A (PDB code: 3EML~\citep{jaakola_2.6_2008}, chain~A) systems were created following the same procedure described above, with the additional inclusion of the following atomic constraints during docking:  
\begin{enumerate}
    \item PPAR$\alpha$: three out of four hydrogen-bond constraints between the R~groups of Ser280, Tyr314, His440, and Tyr464 and the hydrogen-bond acceptor groups of the compounds to be docked must be satisfied,  
    \item ADORA2A: one hydrogen-bond constraint between the R~group of Asn253 and the hydrogen-bond acceptor groups of the ligands,  
    \item D3R: one hydrogen-bond constraint between the R~group of Asp110 and the hydrogen-bond donor group of a protonated nitrogen in the ligand.
\end{enumerate}
The pattern was defined using the SMARTS string: \texttt{"[\#1][\#7+]"}.  

The Glide Standard Precision (SP)~\citep{friesner_extra_2006} protocol was used for docking calculations. Each ligand was docked with the van der Waals radii of atoms having partial charges $\leq 0.15$ scaled by a factor of~0.9. For each ligand, only the best-scoring pose was retained. All calculations were carried out using the OPLS4 force field.

\section{Biological Evaluation}
\label{app: bio-eval}
\subsection{GSK-3$\beta$ kinase assay}
The GSK-3$\beta$ kinase assay was performed as described in~\citep{di_martino_rational_2025}. 
Briefly, the inhibitory potency against human recombinant GSK-3$\beta$ (Carna Biosciences) was evaluated using the LANCE\textsuperscript{\textregistered} Ultra time-resolved fluorescence resonance energy transfer (TR-FRET) assay (PerkinElmer), by measuring phosphorylation of the specific substrate human muscle glycogen synthase (ULight-GS (Ser641/pSer657)), according to the manufacturer’s instructions. 
Test compounds, staurosporine (reference compound), or DMSO (control) were mixed with the enzyme (2~nM) in a buffer containing 50~mM HEPES (pH~7.5), 1~mM EGTA, 10~mM MgCl$_2$, 2~mM DTT, and 0.01\% Tween-20. 
The reaction was initiated by adding 50~nM of the substrate ULight-PASVPPSPSLSRHSSPHQ(pS)ED and 3~$\mu$M ATP, followed by incubation for 1~hour at 23\,$^{\circ}$C.

Following incubation, the reaction was stopped by adding 8~mM EDTA. After 5~min, the anti-phospho-GS antibody labeled with europium chelate was added. After 1 more hour, the kinase reaction was monitored by irradiation at 320~nm, and fluorescence was measured at 615~nm and 665~nm using the EnVision~2014 Multilabel Reader (PerkinElmer). The calculated signal ratio at 665/615~nm was proportional to the extent of ULight-GS phosphorylation. 

Compounds were screened at three concentrations (1, 10, and 50~$\mu$M). 
The most potent compound was further tested in a dose-response format at 11 concentrations ranging from 300~pM to 100~$\mu$M in technical triplicates. 
Results were expressed as percent inhibition of control enzyme activity.

\subsection{Analysis of the Biological Data}
Three-concentration screening and dose-response curves were run in at least three independent experiments, each performed in technical triplicates. 
IC$_{50}$ values were determined by non-linear regression analysis of log[concentration]--response curves, generated from mean replicate values using a four-parameter Hill equation curve fit in GraphPad Prism~8 (GraphPad Software Inc., CA, USA).

Compounds~1, 2, and 3 present high TD values of 0.64 to ChEMBL3652544 (4), 0.61 to ChEMBL250386 (5), and 0.65 to ChEMBL359554 (6), respectively. In addition, we extended the chemical similarity analysis to all kinase inhibitors present in the ChEMBL database (see Supplemental Materials \& Methods section for details). This step was motivated by the conserved nature of kinase binding pockets, where chemically similar scaffolds can often exhibit non-selective interactions. Remarkably, all selected compounds show a minimum TD value above 0.5, even when compared against the full kinase inhibitor set. Specifically, compound~1 presents a minimum TD of 0.51 to ChEMBL3663033 (7), compound~2 to ChEMBL326596 (TD = 0.58, 8), and compound~3 to ChEMBL578061 (TD = 0.63, 9). Representative structures of the most similar GSK-3$\beta$ and kinase inhibitors to our selected compounds are shown in Figure~\ref{fig:bio-activity}A.  

To evaluate the ability of the selected compounds to modulate GSK-3$\beta$ kinase activity, their inhibitory effects were assessed at three concentrations (1, 10, and 50~$\mu$M) using the LANCE$^{\text{\textregistered}}$ Ultra time-resolved fluorescence resonance energy transfer (TR-FRET) assay, as described in the Biological Evaluation section. Compounds~3 and 2 exhibited weak (5~$\pm$~0.7\%) to moderate (49.6~$\pm$~0.9\%) inhibition of GSK-3$\beta$ at the highest tested concentration (50~$\mu$M), respectively. In contrast, compound~1 showed strong inhibition already at 1~$\mu$M and was therefore selected for IC$_{50}$ determination (Table~\ref{tab:bio-assay}). The IC$_{50}$ of compound~1 was calculated to be 314~$\pm$~15.3~nM (Figure~\ref{fig:bio-activity}B).  

\begin{table}[h]
\centering
\caption{Percentage of inhibition of the selected compounds at three different concentrations performed through a TR-FRET assay.}
\label{tab:bio-assay}
\begin{tabular}{lccc}
\toprule
 & 1~$\mu$M & 10~$\mu$M & 50~$\mu$M \\
\midrule
1 & 71.3~$\pm$~1.7 & 88.2~$\pm$~0.1 & 89.9~$\pm$~0.2 \\
2 & 1~$\pm$~3.6   & 12.4~$\pm$~0.8 & 49.6~$\pm$~0.9 \\
3 & n.i. & n.i. & 5~$\pm$~0.7 \\
\bottomrule
\end{tabular}
\end{table}

The 1H-pyrazolo[3,4-b]pyridine moiety of the potent hit compound~1 anchors the molecule to the hinge region of GSK-3$\beta$ through two key hydrogen bonds with the backbone of Asp133 and Val135. Notably, this scaffold is rare among GSK-3$\beta$ inhibitors, with only 19 instances out of 1734 (1.1\%), and similarly underrepresented among all kinase inhibitors in ChEMBL (727 out of 100365, 0.72\%). The high potency of compound~1 may also be explained by an additional hydrogen bond between the nitrogen atom of its 2-methylpyridine ring and the side-chain amino group of Lys85, a conserved residue known to enhance inhibitor affinity. These findings highlight the model’s ability to generate chemically novel yet functionally relevant hits. Despite not being explicitly fine-tuned on GSK-3$\beta$ ligands, MolRNN successfully designed a potent hit by leveraging general kinase features embedded in the REINVENT training dataset. This is further supported by the t-SNE analysis (Figure~\ref{fig:bio-activity}C), where compound~1 (red star), although structurally novel, clusters within the dense chemical space of known GSK-3$\beta$ binders (gold dots).  

In light of its strong inhibitory activity, we benchmarked compound~1 against known GSK-3$\beta$ inhibitors to better assess its relative potency. Specifically, we compared its activity value, expressed as --log(activity), with those of known GSK-3$\beta$ inhibitors. As shown in the violin plot (Figure~\ref{fig:bio-activity}D), compound~1 (red star) outperforms both the full ligand set (light blue shape) and the subset of hit-like GSK-3$\beta$ inhibitors (blue dots), with a --log(activity) value of 6.50 compared to median values of 6.39 and 5.89, respectively.

\begin{figure}[htbp]
    \centering
    \includegraphics[width=1.0\linewidth]{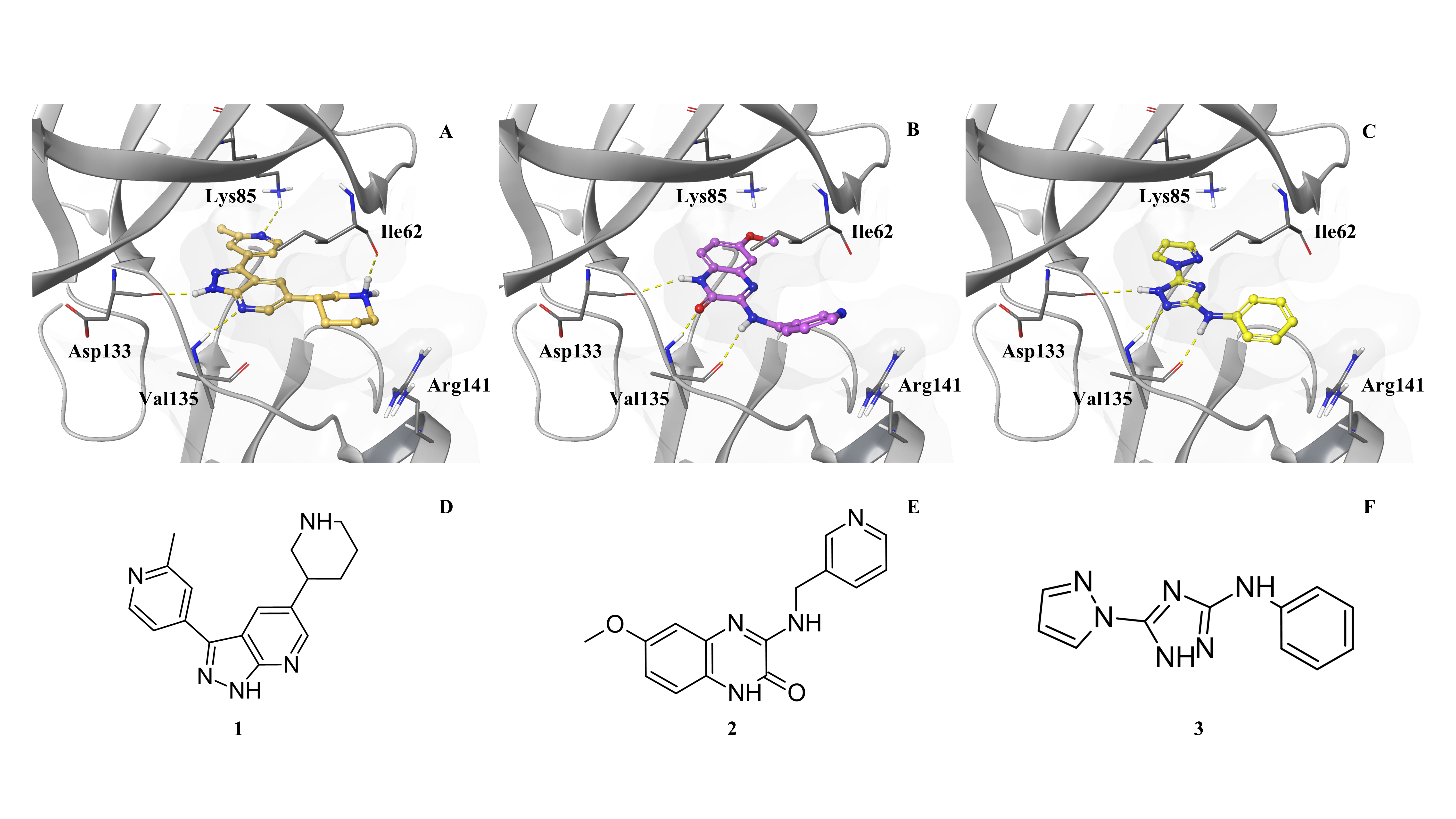} 
    \caption{Predicted binding conformations of compounds~1 (carbon atoms in gold, A), 2 (carbon atoms in purple, B), and 3 (carbon atoms in yellow, C) at the binding site of GSK-3$\beta$. In panels~A--C, the protein structure is shown as a thin grey ribbon. Residues interacting with the docked compound are displayed in stick representation with light grey carbons and explicitly labelled. Hydrogen bonds are depicted as yellow dashed lines, while a grey mesh highlights the boundaries of the binding pocket within 5~\AA\ of each ligand. The chemical structures of compounds~1, 2, and 3 are shown in panels~D--F, respectively.}
    \label{fig:selected-compounds}
\end{figure}

\newpage
\section{Per-target Docking Results and Analysis and Physicochemical Profiles}
\label{app:per-target-kl}

\begin{table}[htbp]
  \centering
  \caption{KL divergence between the docking score distributions of generated molecules and target ligand sets for each of the seven investigated targets. Lower values indicate higher similarity between generated and reference distributions.}
  \label{tab:kl-by-target}
  \begin{tabular}{lc}
    \toprule
    Target & KL divergence \\
    \midrule
    ADORA2A      & $0.027 \pm 0.015$ \\
    D3R          & $0.055 \pm 0.040$ \\
    GSK-3$\beta$ & $0.019 \pm 0.012$ \\
    HSP90$\alpha$& $0.039 \pm 0.030$ \\
    PPAR$\alpha$ & $0.098 \pm 0.074$ \\
    SRC          & $0.122 \pm 0.089$ \\
    Thrombin     & $0.084 \pm 0.063$ \\
    \bottomrule
  \end{tabular}
\end{table}

\begin{table*}[htbp]
  \centering
  \caption{For each target, the median docking scores of the full and Hit-like ligand sets, along with the percentage of generated molecules (across all trained models) scoring better than each respective median.}
  \label{tab:medians-full-vs-hitlike}
  \scriptsize
  \setlength{\tabcolsep}{4pt}
  \renewcommand{\arraystretch}{1.05}
  \begin{tabular}{l l r r r r}
    \toprule
    \textbf{Experiment} & \textbf{Target} & 
    \textbf{Median (Full)} & \textbf{Median (Hit-like)} &
    \textbf{\% Better (Full)} & \textbf{\% Better (Hit-like)} \\
    \midrule
    DiGress (Hit-like)   & ADORA2A  & $-8.89$ & $-8.86$ & 1.63 & 1.79 \\
                         & PPAR$\alpha$  & $-9.15$ & $-6.92$ & 0.38 & 23.28 \\
                         & HSP90$\alpha$ & $-6.03$ & $-6.55$ & 39.97 & 17.78 \\
                         & GSK-3$\beta$  & $-7.27$ & $-7.83$ & 10.13 & 3.26 \\
                         & Thrombin      & $-7.59$ & $-6.10$ & 0.98  & 35.38 \\
                         & SRC           & $-7.71$ & $-7.55$ & 5.23  & 7.34 \\
                         & D3R           & $-6.81$ & $-6.46$ & 13.34 & 23.61 \\
    \midrule
    DiGress (REINVENT)   & ADORA2A  & $-8.89$ & $-8.86$ & 2.12 & 2.29 \\
                         & PPAR$\alpha$  & $-9.15$ & $-6.92$ & 0.35 & 21.42 \\
                         & HSP90$\alpha$ & $-6.03$ & $-6.55$ & 42.69 & 19.74 \\
                         & GSK-3$\beta$  & $-7.27$ & $-7.83$ & 12.18 & 4.38 \\
                         & Thrombin      & $-7.59$ & $-6.10$ & 1.22 & 40.20 \\
                         & SRC           & $-7.71$ & $-7.55$ & 6.74 & 9.32 \\
                         & D3R           & $-6.81$ & $-6.46$ & 12.45 & 21.72 \\
    \midrule
    DiGress (Hit fine-tune) & ADORA2A  & $-8.89$ & $-8.86$ & 2.23 & 2.47 \\
                         & PPAR$\alpha$  & $-9.15$ & $-6.92$ & 0.54 & 25.34 \\
                         & HSP90$\alpha$ & $-6.03$ & $-6.55$ & 40.22 & 17.57 \\
                         & GSK-3$\beta$  & $-7.27$ & $-7.83$ & 11.60 & 3.94 \\
                         & Thrombin      & $-7.59$ & $-6.10$ & 1.31 & 41.16 \\
                         & SRC           & $-7.71$ & $-7.55$ & 6.44 & 8.96 \\
                         & D3R           & $-6.81$ & $-6.46$ & 14.12 & 24.44 \\
    \midrule
    MolRNN (Hit-like)    & ADORA2A  & $-8.89$ & $-8.86$ & 2.07 & 2.27 \\
                         & PPAR$\alpha$  & $-9.15$ & $-6.92$ & 0.38 & 22.86 \\
                         & HSP90$\alpha$ & $-6.03$ & $-6.55$ & 42.07 & 19.50 \\
                         & GSK-3$\beta$  & $-7.27$ & $-7.83$ & 15.58 & 6.35 \\
                         & Thrombin      & $-7.59$ & $-6.10$ & 1.16 & 36.95 \\
                         & SRC           & $-7.71$ & $-7.55$ & 7.01 & 9.71 \\
                         & D3R           & $-6.81$ & $-6.46$ & 13.05 & 22.53 \\
    \midrule
    MolRNN (REINVENT)    & ADORA2A  & $-8.89$ & $-8.86$ & 2.64 & 2.86 \\
                         & PPAR$\alpha$  & $-9.15$ & $-6.92$ & 0.46 & 22.40 \\
                         & HSP90$\alpha$ & $-6.03$ & $-6.55$ & 43.05 & 20.32 \\
                         & GSK-3$\beta$  & $-7.27$ & $-7.83$ & 15.60 & 6.39 \\
                         & Thrombin      & $-7.59$ & $-6.10$ & 1.52 & 41.89 \\
                         & SRC           & $-7.71$ & $-7.55$ & 7.75 & 10.62 \\
                         & D3R           & $-6.81$ & $-6.46$ & 13.48 & 22.81 \\
    \midrule
    MolRNN (Hit fine-tune) & ADORA2A  & $-8.89$ & $-8.86$ & 2.90 & 3.17 \\
                         & PPAR$\alpha$  & $-9.15$ & $-6.92$ & 0.52 & 24.91 \\
                         & HSP90$\alpha$ & $-6.03$ & $-6.55$ & 51.82 & 25.16 \\
                         & GSK-3$\beta$  & $-7.27$ & $-7.83$ & 20.59 & 8.83 \\
                         & Thrombin      & $-7.59$ & $-6.10$ & 1.55 & 45.29 \\
                         & SRC           & $-7.71$ & $-7.55$ & 9.46 & 12.89 \\
                         & D3R           & $-6.81$ & $-6.46$ & 15.89 & 27.28 \\
    \midrule
    GraphINVENT (Hit-like) & ADORA2A  & $-8.89$ & $-8.86$ & 2.15 & 2.35 \\
                         & PPAR$\alpha$  & $-9.15$ & $-6.92$ & 0.38 & 22.47 \\
                         & HSP90$\alpha$ & $-6.03$ & $-6.55$ & 38.35 & 17.13 \\
                         & GSK-3$\beta$  & $-7.27$ & $-7.83$ & 14.00 & 5.44 \\
                         & Thrombin      & $-7.59$ & $-6.10$ & 1.36 & 39.10 \\
                         & SRC           & $-7.71$ & $-7.55$ & 6.67 & 9.33 \\
                         & D3R           & $-6.81$ & $-6.46$ & 12.45 & 22.25 \\
    \midrule
    GraphINVENT (REINVENT) & ADORA2A  & $-8.89$ & $-8.86$ & 1.95 & 2.13 \\
                         & PPAR$\alpha$  & $-9.15$ & $-6.92$ & 0.28 & 19.37 \\
                         & HSP90$\alpha$ & $-6.03$ & $-6.55$ & 39.78 & 17.95 \\
                         & GSK-3$\beta$  & $-7.27$ & $-7.83$ & 13.38 & 5.38 \\
                         & Thrombin      & $-7.59$ & $-6.10$ & 1.19 & 37.98 \\
                         & SRC           & $-7.71$ & $-7.55$ & 6.43 & 8.90 \\
                         & D3R           & $-6.81$ & $-6.46$ & 11.08 & 20.15 \\
    \midrule
    GraphINVENT (Hit fine-tune) & ADORA2A  & $-8.89$ & $-8.86$ & 2.34 & 2.54 \\
                         & PPAR$\alpha$  & $-9.15$ & $-6.92$ & 0.34 & 21.21 \\
                         & HSP90$\alpha$ & $-6.03$ & $-6.55$ & 39.07 & 17.55 \\
                         & GSK-3$\beta$  & $-7.27$ & $-7.83$ & 14.86 & 6.04 \\
                         & Thrombin      & $-7.59$ & $-6.10$ & 1.42 & 39.79 \\
                         & SRC           & $-7.71$ & $-7.55$ & 7.18 & 9.93 \\
                         & D3R           & $-6.81$ & $-6.46$ & 11.92 & 20.97 \\
    \bottomrule
  \end{tabular}
\end{table*}

\begin{figure}[htbp]
    \centering
    \includegraphics[width=0.7\linewidth]{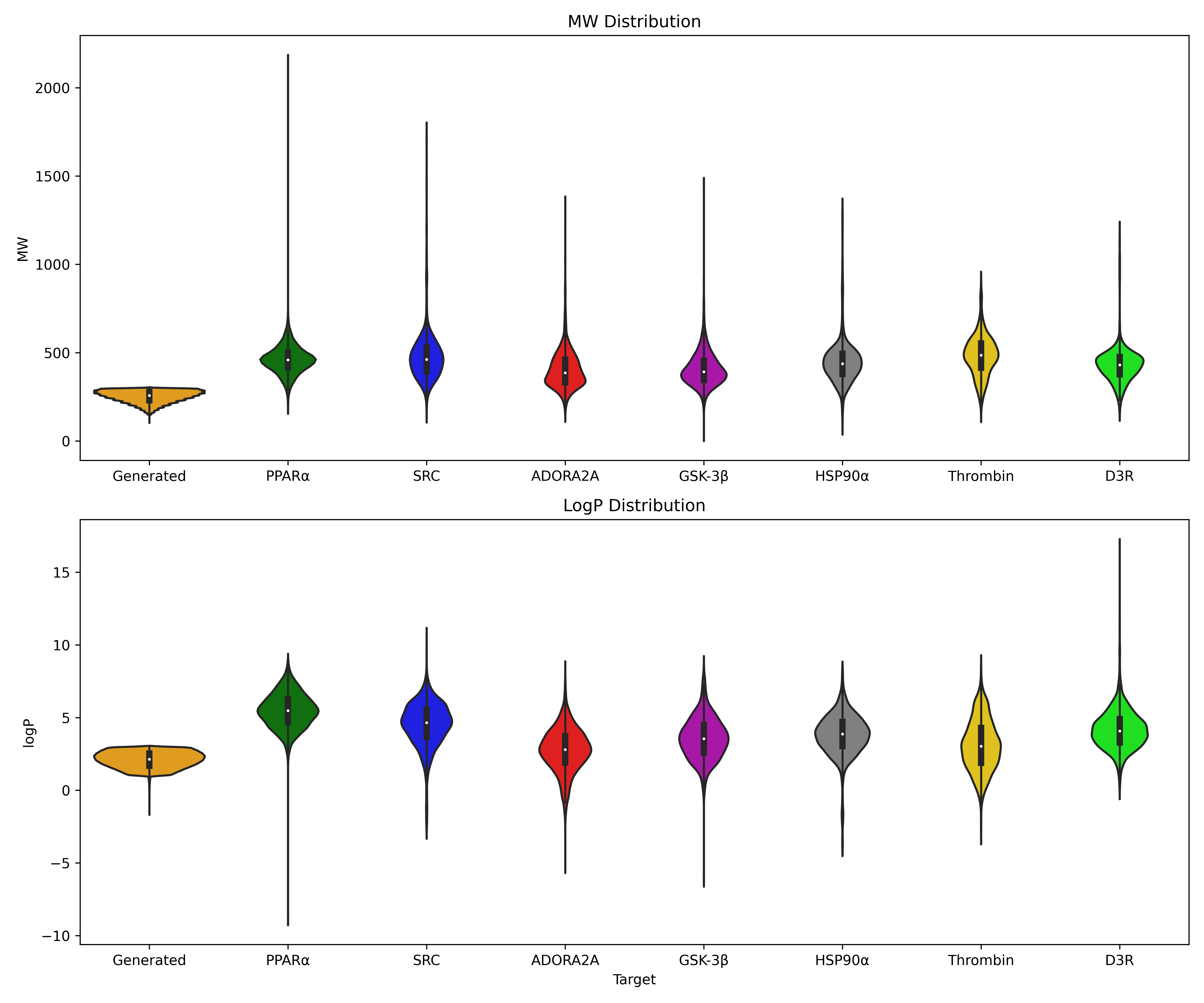} 
    \caption{Comparison of MW (upper panel) and logP (lower panel) of the generated molecules with respect to each target ligand set.}
    \label{fig:violin_plots_noft_set}
\end{figure}
\end{document}